\documentclass[letterpaper, 10 pt, journal, twoside]{IEEEtran}
\usepackage{amsmath,amsfonts}
\usepackage{etoolbox} 
\usepackage{amssymb}
\usepackage{array}
\usepackage[caption=false,font=normalsize,labelfont=sf,textfont=sf]{subfig}
\usepackage{bm}
\usepackage{textcomp}
\usepackage{stfloats}
\usepackage{xcolor}
\usepackage[normalem]{ulem}
\usepackage[hidelinks]{hyperref}
\usepackage{url}
\usepackage{verbatim}
\usepackage{graphicx}
\usepackage{cite}
\usepackage{booktabs}
\usepackage{makecell}
\usepackage{microtype}
\usepackage{placeins}
\AtBeginEnvironment{equation}{\small}
\AtBeginEnvironment{align}{\small}
\AtBeginEnvironment{gather}{\small}
\AtBeginEnvironment{multline}{\small}

\begin{document}

\title{Ask to Land: Learning a Mask-Conditioned Open-Vocabulary Landing Policy via Differentiable Physics}

\author{Fangyu Sun,  Guanyan Wu, Yang Deng, Jiahao Cui, Fanxing Li, Yuxiang Huang, Wenxian Yu, Danping Zou
\thanks{Manuscript received: May 19, 2026; Revised: July 19, 2026; Accepted: September 10, 2026. This paper was recommended for publication by Editor Giuseppe Loianno upon evaluation of the Associate Editor and Reviewers' comments. This work was supported by the National Science Foundation of China (62073214).\textit{(Corresponding author: Danping Zou.)}}
\thanks{All authors are with the School of Automation and Perception, Shanghai Jiao Tong University, China (e-mail: dpzou@sjtu.edu.cn).}
\thanks{Digital Object Identifier (DOI): see top of this page.}}

\markboth{IEEE Robotics and Automation Letters. Preprint Version. ACCEPTED SEPTEMBER, 2026}
{Sun \MakeLowercase{\textit{et al.}}: Ask to Land: Learning a Mask-Conditioned Open-Vocabulary Landing Policy via Differentiable Physics} 

\maketitle

\begin{abstract}
Existing drone landing systems, whether model-based or learning-based, typically rely on predefined visual markers (e.g., ArUco tags) or are trained for specific platform categories, and often assume structured target motion. Recent open-vocabulary studies enable text-prompted target selection, but integrating this capability with learned closed-loop landing control remains underexplored. To this end, we present \textit{Ask to Land}, a mask-conditioned framework that enables a quadrotor to land on various platforms described by a user in natural language, including randomly moving platforms. The core of our approach is the integration of a frozen state-of-the-art open-vocabulary perception frontend (YOLO-World with MobileSAM) and a control policy learned by differentiable physics. By providing real-time, task-agnostic semantic segmentation masks as policy inputs, our framework enables the drone to land on prompt-selected targets without target-specific fine-tuning while continuously tracking dynamic target motion during descent. The frozen frontend processes the prompt, while the policy receives the mask without text embeddings or category labels. To efficiently optimize the control policy, we leverage differentiable physics learning via Backpropagation-Through-Time (BPTT), and employ a carefully designed reward that enforces visual alignment with the landing target and altitude-adaptive descent speed, together with stable control. Real-world deployments validate that our approach achieves landings on diverse user-specified objects, including static and moving platforms.
\end{abstract}


\section{Introduction}

Imagine a rescue drone deployed in a disaster zone. Its mission is to land safely on a moving pickup truck or a distinctly colored supply box. The drone does not know the exact appearance of the target beforehand, and the target's position changes unpredictably. Landing in such unstructured scenarios requires the drone to recognize user-specified targets on the fly and continuously adjust its descent speed to match unknown target motions.

Current autonomous landing systems struggle with these real-world conditions. Classical model-based methods often require predefined visual markers (like ArUco tags) or exact geometric priors of the landing pad \cite{ghommam2017moving,ibvs_moving_target_2016,falanga2017moving,keipour2022visual}. Learning-based methods typically train the drone for specific object categories \cite{landing2010,21iros_landing,iros18_moving,ladosz2024autonomous,landing2025}. Recent open-vocabulary systems enable prompt-based landing-target selection, but do not learn a closed-loop policy that maps the selected target representation directly to low-level commands throughout the complete landing process \cite{bong2023dovesei,bong2023peace,bong2024blabberseg,cai2025llmland}. Furthermore, existing dynamic landing systems often assume the target moves along a known path \cite{iros18_moving,ladosz2024autonomous}, and this assumption frequently fails when a platform changes its trajectory randomly.

\begin{figure}[!t]
    \centering
    \includegraphics[width=\linewidth]{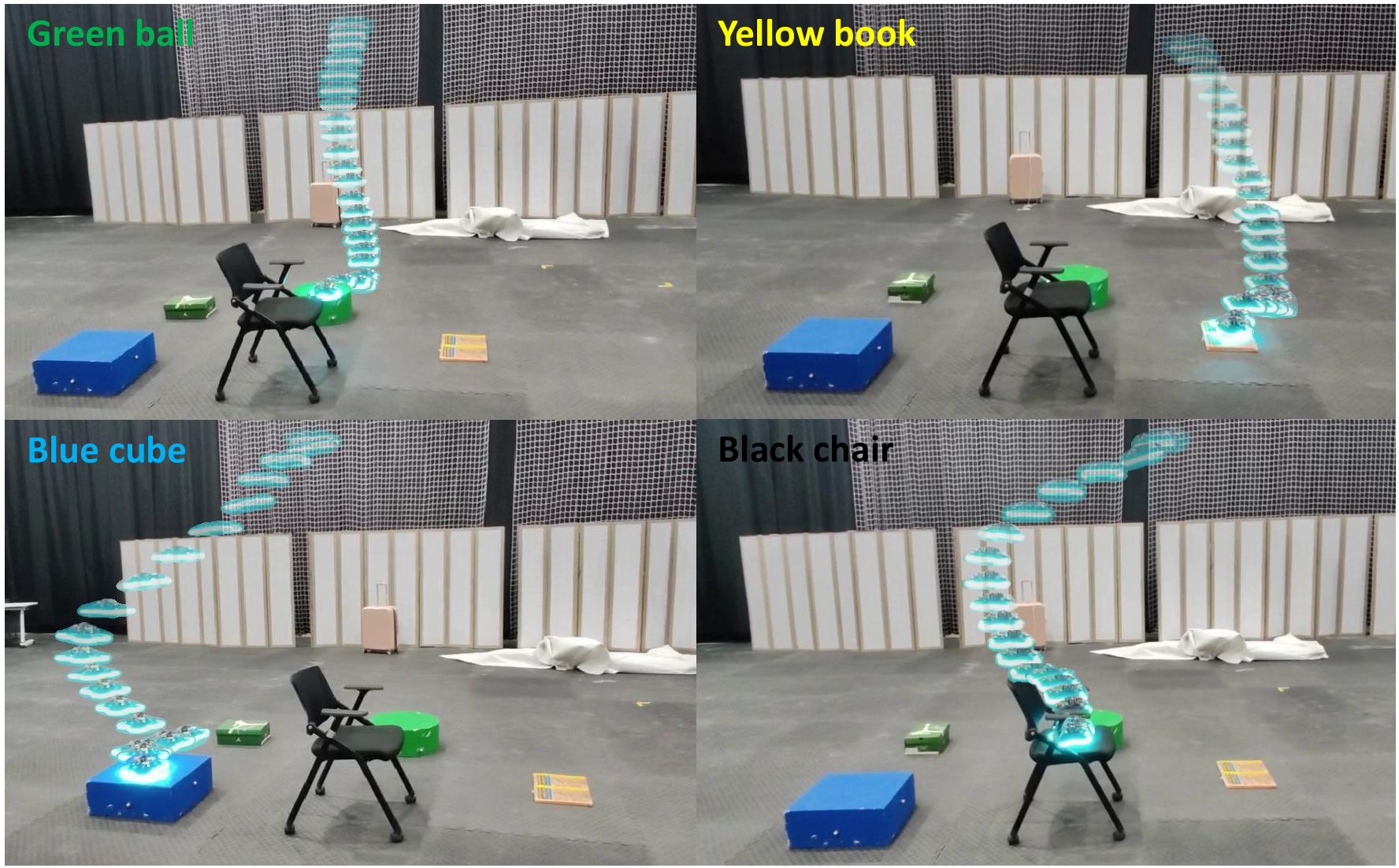}
    \caption{Open-vocabulary multi-platform landings. The frozen frontend selects prompt-specified platforms for the mask-conditioned policy without target-specific fine-tuning.}
    \label{fig:first_fig_static}
\end{figure}

To address these challenges, we propose \textit{Ask to Land}, a mask-conditioned framework designed to enable robust autonomous drone landing on diverse, user-specified platforms, including dynamic platforms that move along random trajectories. First, to overcome the limitations of predefined platform categories and achieve prompt-based target selection and controller transfer, we incorporate the semantic segmentation mask as the policy input. During real-world deployment, we leverage a frozen open-vocabulary detector, YOLO-World \cite{yolo-world}, in conjunction with a lightweight segmentation module, MobileSAM \cite{mobile_sam}. Guided by natural language prompts, this frontend generates real-time, task-agnostic semantic masks of the desired landing surface. By conditioning the control policy on these semantic representations rather than platform-specific labels or geometric priors, the learned policy can be applied to user-specified platforms of varying shapes, sizes, and categories, without requiring target-specific fine-tuning.

Beyond open-vocabulary perception, we tackle the dynamic landing problem with a differentiable composite reward that first enforces persistent semantic visual alignment with the moving target, and then jointly encourages planar tracking and altitude-adaptive descent speed. In addition, we employ domain randomization and curriculum learning during training. We randomize platform geometry and appearance, ground-plane motion with varying low speeds, as well as the initial positions of both the platform and the drone. The curriculum gradually increases task difficulty, starting from static platforms and progressing toward dynamic tasks with random trajectories and speeds. Finally, to improve training efficiency, we optimize the policy with Backpropagation-Through-Time (BPTT) through fully differentiable quadrotor dynamics, exploiting analytic gradients through the dynamics for faster convergence.

Our main contributions are:

{\sloppy
\begin{itemize}
    \item We introduce a mask-conditioned open-vocabulary landing system that integrates prompt-based target selection with a landing policy trained via differentiable physics through a binary-mask interface. The policy maps the mask directly to control commands without receiving language embeddings or category labels.
    \item We design an appropriate differentiable reward function that couples persistent visual alignment with altitude-adaptive descent, enabling stable tracking of platforms with random trajectories and varying speeds.
    \item We validate our system on a real quadrotor using prompt-selected static and moving platforms. The system achieves accurate and stable landings across diverse targets while adapting its descent behavior throughout the approach. 
\end{itemize}
}

\begin{figure*}[htbp]
    \centering
    \includegraphics[width=0.95\linewidth]{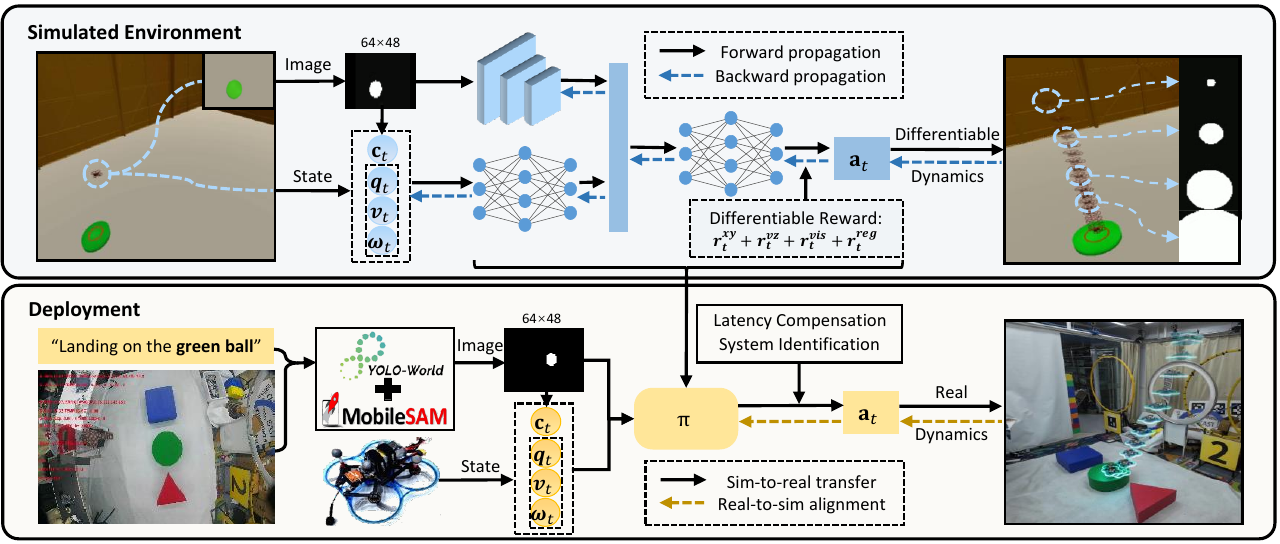}
    \caption{System overview. BPTT trains the mask-conditioned policy through differentiable quadrotor dynamics, while the mask is used only in the forward pass. At deployment, the frozen YOLO-World--MobileSAM frontend converts the prompted target into the binary mask.}
    \label{fig:system}
\end{figure*}

\section{Related Works}
Autonomous drone landing typically assumes either fixed platforms or constrained target motion. Traditional model-based approaches, such as visual servoing, rely on pre-defined visual markers or geometric priors~\cite{ibvs_moving_target_2016,falanga2017moving,ghommam2017moving,keipour2022visual}. Learning-based methods can reduce manual design but are still evaluated on fixed platform categories or dedicated landing targets~\cite{landing2010,21iros_landing,iros18_moving,ladosz2024autonomous,landing2025}. For moving targets, prior work often assumes predefined trajectories or structured motion patterns~\cite{iros18_moving,ladosz2024autonomous}. Recent RL work achieves fast random landing but still relies on a specific marker~\cite{shin2026uncertain_motion}. In contrast, our setting considers prompt-selected landing targets and random trajectories with randomly varying speeds during descent. Open-vocabulary perception has recently become feasible through Vision-Language Models and foundation models such as SAM~\cite{clip,sam}. Previous works in robotic manipulation and navigation leverage language-conditioned visual representations to achieve zero-shot semantic generalization~\cite{voxposer,lmnav,rt2}. Specifically, DOVESEI uses prompted segmentation with reactive visual servoing~\cite{bong2023dovesei}, PEACE refines segmentation prompts~\cite{bong2023peace}, BlabberSeg enables efficient onboard aerial segmentation~\cite{bong2024blabberseg}, and LLM-Land combines language reasoning with MPC~\cite{cai2025llmland}. Drawing inspiration from these methods, we combine YOLO-World and MobileSAM as a lightweight semantic frontend to enable flexible recognition and segmentation of prompt-selected landing targets. Our distinction lies in learning a mask-to-action closed-loop policy by backpropagating through a differentiable model of the full quadrotor dynamics, and in evaluating the controller over the complete landing process on prompt-selected static objects and randomly moving platforms.

Differentiable physics learning offers an efficient route for optimizing control policies by propagating analytic gradients through physical dynamics \cite{loco1,loco2,shac}. In quadrotor research, it has shown strong sample efficiency in tasks such as swarm obstacle avoidance \cite{yuang_nmi}, vision-based stabilization \cite{diff2}, and optical flow-based agile flight \cite{hu}. Nevertheless, these works primarily focus on aggressive flight, rather than stable grounded landing tasks for static and moving targets. Building on these advances, we use differentiable physics learning to learn a mask-conditioned landing policy that couples target tracking and altitude-adaptive descent in a unified framework.

\section{Methodology}
The overall architecture of our \textit{Ask to Land} system is illustrated in Fig.~\ref{fig:system}. In the following subsections, we detail the core components of our methodology: the real-time open-vocabulary perception frontend, the problem formulation via differentiable physics learning, and the training details.
\subsection{Real-time Open-Vocabulary Perception}
\begin{figure}[htbp]
    \vspace{-5pt}
    \centering
    \includegraphics[width=0.88\linewidth]{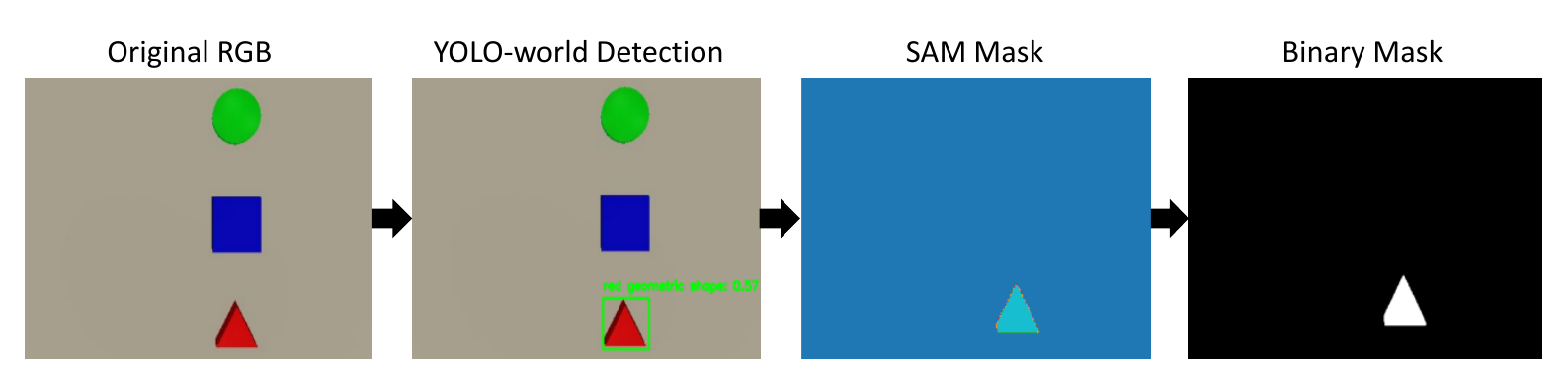}
    \caption{Real-world semantic mask generation pipeline.}
    \label{fig:semantic_mask_pipeline}
    \vspace{-6pt}
\end{figure}

For prompt-based target selection, we use a perception frontend that leverages the synergy between open-vocabulary object detection and promptable semantic segmentation. Unlike conventional closed-set visual pipelines that rely on predefined category labels, our approach formulates target identification as a language-conditioned perception task. By conditioning the detection module on arbitrary linguistic descriptions (e.g., ``book'' or ``black chair''), the system can instantly adapt to diverse platform semantics and geometries without task-specific fine-tuning. As shown in Fig.~\ref{fig:semantic_mask_pipeline}, the perception pipeline operates sequentially through three primary stages:

\begin{itemize}
    \item \textbf{Text-Conditioned Target Proposal:} Given a streaming monocular RGB image and a user-specified text prompt, the open-vocabulary detector, YOLO-World \cite{yolo-world}, generates a set of candidate bounding boxes. To ensure robust target acquisition during dynamic flight, we employ a relaxed confidence threshold to maximize recall, followed by a spatial-confidence heuristic that isolates the optimal target bounding box exhibiting the highest probability score and minimal spatial deviation from the image principal point.
    \item \textbf{Prompt-Driven Mask Generation:} The selected bounding box is subsequently utilized as a spatial prior for MobileSAM \cite{mobile_sam}. By extracting geometric prompts from the bounding box, MobileSAM selectively attends to the specified Region of Interest (RoI), yielding a high-fidelity, instance-level binary segmentation mask. This cascaded coupling effectively isolates the landing target while filtering out unstructured background clutter.
    \item \textbf{Semantic State Representation:} To satisfy the dimensional constraints of the control policy and minimize computational latency, the high-resolution binary mask is downsampled via bilinear interpolation and subsequently binarized to formulate a compact semantic observation matrix $\bm{I}_t \in \mathbb{R}^{64 \times 48}$. This task-agnostic representation intrinsically decouples the control policy from the visual appearance of the platform, thereby supporting transfer across the mask geometries, apparent scales, and visibility conditions represented during evaluation.
\end{itemize}

\subsection{Problem Formulation via Differentiable Physics Learning}
The open-vocabulary landing task is formulated as a continuous control problem where the objective is to find an optimal policy $\pi_\theta$ that safely guides a quadrotor to land on an arbitrary platform specified by a natural language prompt $\mathcal{P}$. At each timestep $t$, the state is $s_t \in \mathcal{S}$, the control input is $a_t \in \mathcal{A}$, and the agent receives a reward $r_t = r(s_t, a_t)$. An observation $o_t \in \mathcal{O}$ is generated based on the current state $s_t$ through an observation model $h:\mathcal{S} \rightarrow \mathcal{O}$, such that $o_t = h(s_t)$. The system dynamics are governed by a function $f:\mathcal{S} \times \mathcal{A} \rightarrow \mathcal{S}$, which describes the forward process of the system $s_{t+1} = f(s_t, a_t)$. The mask-conditioned control policy directly maps the observation to the control input $a_t = \pi_\theta(o_t)$.

The objective function $\mathcal{J}(\theta)$ represents the cumulative reward, which is maximized using gradient ascent to determine the optimal parameters $\theta^\star$:
\begin{equation}
    \theta^\star = \arg\max_\theta \mathcal{J}(\theta),
\end{equation}
\begin{align}
    \mathcal{J}(\theta) &= \sum_{t=0}^{N-1} \gamma^t r_t(s_t, a_t) = \sum_{t=0}^{N-1} \gamma^t r_t(s_t, \pi_\theta(o_t)) \nonumber \\
    &= \sum_{t=0}^{N-1} \gamma^t r_t(s_t, \pi_\theta(h(s_t))).
\end{align}
\begin{equation}
    \theta_{t+1} \leftarrow \theta_t + \alpha \nabla_\theta \mathcal{J}(\theta_t),
\end{equation}
where $\gamma$ is the discount factor, and $\alpha$ is the learning rate. Since the system dynamics and the reward are differentiable, this objective can be directly optimized via analytic gradient ascent with respect to $\theta$.

\textbf{Backpropagation-Through-Time (BPTT):}
By exploiting the temporal structure of $\mathcal{J}(\theta)$ and applying the chain rule along the time axis, the policy gradient can be obtained via BPTT as
\begin{equation}\label{eq:shac_gradient}
\begin{aligned}
    &\nabla_\theta \mathcal{J}(\theta) = \\
    &\sum_{t=0}^{N-1} \gamma^t \sum_{i=0}^{t-1} \frac{\partial {r}_t}{\partial s_t}
    \left( \prod_{j=i+1}^{t-1} \underbrace{\left( \frac{\partial s_{j+1}}{\partial s_j} \right)}_{\text{differentiable dynamics}} \right)
    \frac{\partial s_{i+1}}{\partial a_i}\frac{\partial a_i}{\partial \theta} \\
    &+ \sum_{t=0}^{N-1} \gamma^t \frac{\partial {r}_t}{\partial a_t} \frac{\partial a_t}{\partial \theta}
\end{aligned}
\end{equation}
where the matrix of partial derivatives ${\partial s_j}/{\partial s_{j-1}}$ is the Jacobian of the quadrotor dynamics. In practical implementation, this continuous process is discretized into a time-stepped evolution of the system, expressed as $s_{t+1} = s_t + \dot{s}_t dt$. During backpropagation, gradients are propagate through the continuous observation of $s_t$ and the differentiable dynamics, whereas the inputs derived from the mask branch of $h(s_t)$ are treated as fixed and detached from the computational graph. 

\subsection{Differentiable Dynamics}
We employ a fully differentiable quadrotor dynamics model $\dot{\bm{s}_t}$ for backpropagation in Eq.~\ref{dynamic}. 
\begin{equation}\label{dynamic}
    \dot{\bm{s}} = 
    \begin{bmatrix}
    \dot{\bm{p}}_W \\
    \dot{\bm{q}} \\
    \dot{\bm{v}}_{W} \\
    \dot{\bm{\omega}}
    \end{bmatrix}
    =
    \begin{bmatrix}
    \bm{v}_{W} \\
    \bm{q} \cdot 
    \begin{bmatrix}
    0 \\
    \bm{\omega}/2
    \end{bmatrix} \\
    \frac{1}{m} \left( \bm{R}_{WB} \left( \bm{f}_{T} + \bm{f}_{D} \right) \right) + \bm{g} \\
    \bm{J}^{-1} \left( \bm{\tau}_{T} + \bm{\tau}_{D} - \bm{\omega} \times \bm{J} \bm{\omega} \right) \\
    \end{bmatrix}
\end{equation}
where $m$ and $\bm{J}$ are the mass and diagonal moment-of-inertia matrix, $\bm{p}_W$ and $\bm{v}_W$ are the position and velocity vectors in the world frame, unit quaternions $\bm{q}$ are used to represent orientations, $\bm{R}_{WB}$ is the rotation matrix from the body frame to the world frame, $\boldsymbol{\omega}$ represents angular velocity in the body frame. $\bm{g}$ denotes the gravity vector. $\bm{f}_T$ and $\bm{\tau}_T$ denote the collective thrust in the body-$z$ axis and cumulative body torques produced by the rotors. $\bm{f}_D$ and $\bm{\tau}_D$ denote aerodynamic drag force and torque.

\subsection{Training Details}
\subsubsection{Observation Space}
The observation space is defined by $\bm{o}_t = \{\bm{I}_t, \bm{c}_t, \bm{s}_t\}$, where $\bm{I}_t \in \mathbb{R}^{64 \times 48}$ is the downsampled binary semantic mask, $\bm{s}_t = \left[\bm{q}_{t}, \bm{v}_t, \bm{\omega}_t \right] \in \mathbb{R}^{10}$ is the proprioceptive state vector and $\bm{c}_t = [x_t, y_t, \mathbb{I}_{vis}] \in \mathbb{R}^{3}$ is the visual alignment vector extracted from the semantic mask. $x_t$ and $y_t$ represent the normalized horizontal and vertical coordinates of the landing platform's centroid in the image plane, and $\mathbb{I}_{vis} \in \{0,1\}$ indicates the visibility of the platform. During simulation, the rendered semantic segmentation image is obtained directly from the simulator. Pixels belonging to the landing-platform IDs are set to one, while background pixels are set to zero. The mask-derived measurements are:
\begin{align}
x_t &= \frac{x_{\text{center}} - W/2}{W/2}, \label{eq:u_t} \\
y_t &= \frac{y_{\text{center}} - H/2}{H/2}. \label{eq:v_t}
\end{align}
where $(x_{\text{center}}, y_{\text{center}})$ denotes the centroid coordinates of the segmented platform region in the image of dimensions $H \times W$. $\bm{q}_{t} \in \mathbb{R}^{4}$ denotes the orientation, $\bm{v}_t \in \mathbb{R}^{3}$ and $\bm{\omega}_t \in \mathbb{R}^{3}$ represent the linear velocity and angular velocity in the body frame, respectively. It is worth noting that the policy does not require any prior information (position, depth, or manual markers) about the landing platform and relies solely on the visual input.

\subsubsection{Action Space}
Due to its stability in previous quadrotor tasks \cite{CTBR1, CTBR2, CTBR3}, our mask-conditioned policy is trained to map observations directly to collective thrust and body rate (CTBR) commands. The action is defined by $\mathbf{a}_t=[T,\omega_x,\omega_y,\omega_z]\in \mathbb{R}^{4}$. These commands are then mapped to the individual motor speeds by a low-level controller.

\subsubsection{Reward Function}
We define a differentiable reward function to guide the quadrotor toward a stable landing. At each timestep $t$, the total differentiable reward $r_t$ is computed as a weighted sum of four terms:
\begin{equation}
    r_t = r_t^{xy} + r_t^{vz} + r_t^{vis} + r_t^{reg}
    \label{eq:total_reward}
\end{equation}
\noindent where $r_t^{xy}$ denotes the horizontal progress reward, $r_t^{vz}$ is the adaptive descent velocity reward, $r_t^{vis}$ represents the visual alignment reward, and $r_t^{reg}$ comprises regularization penalties for stability.

\textbf{Horizontal Progress Reward:} To facilitate precise positioning over the landing target, we define a distance-based reward in the planar axes:
\begin{equation}
    r_t^{xy} = -\lambda_{xy} \| \mathbf{p}_{t}^{xy} - \mathbf{g}_{t}^{xy} \|
\end{equation}
\noindent where $\mathbf{p}_{t}^{xy}$ and $\mathbf{g}_{t}^{xy}$ denote the position of the quadrotor and the (possibly moving) landing platform in the x-y plane at time $t$, respectively. 

\textbf{Adaptive Descent Reward:} Unlike standard positioning tasks, stable landing requires a velocity profile that decreases as altitude diminishes. We design an adaptive velocity reward $r_t^{vz}$ that encourages the drone to track a height-dependent reference velocity $v_{ref}$:
\begin{equation}
    r_t^{vz} = \frac{\lambda_z}{1 + |v_{t}^{z} + v_{ref}|}
\end{equation}
\noindent where $v_{t}^{z}$ is the current vertical velocity (negative during descent). The reference velocity is defined continuously as a function of the current altitude $h_t$:
\begin{equation}
    v_{ref} = \text{clip}(\beta \cdot h_t, v_{min}, v_{max})
\end{equation}
\noindent Here, $\beta$ is the velocity coefficient, which controls the reference descent speed. The descent speed is clamped between $v_{min}$ and $v_{max}$. This reward design ensures that the descent speed adapts automatically with altitude, preventing both excessively aggressive and stalled behaviors, and plays a crucial role in producing a smooth and continuous landing trajectory.

\textbf{Visual Alignment Reward:} To ensure the drone keeps the landing platform within its field of view and to explicitly encourage the policy to learn to land from the visual input, we introduce a visual reward $r_t^{vis}$ defined from the analytically projected target centroid:

\begin{equation}
r_t^{vis} =
\begin{cases}
\lambda_{vis}\left(1-\|\mathbf{c}_t^{proj}\|\right), & \mathbb{I}_{vis}=1,\\
0, & \mathbb{I}_{vis}=0.
\end{cases}
\end{equation}

\noindent where $\mathbf{c}_t^{proj}\in\mathbb{R}^2$ is the normalized image-plane coordinate of the target centroid obtained by analytic camera projection. When the mask is in view, $r_t^{vis}$ reduces to the smooth term $(1-\|\mathbf{c}_t^{proj}\|)$ whose gradient with respect to the quadrotor state is defined through the camera projection. When the mask is out of view, $r_t^{vis}$ and its gradient are set to zero, and $r_t^{xy}$ guides the drone back toward the platform until visibility. This term encourages the agent to center the mask in its field of view whenever possible, which enables our policy to achieve both static and dynamic landings.

\textbf{Stability and Regularization:} To minimize oscillation and ensure smooth control, we incorporate regularization terms for orientation, angular velocity, yaw angle, and control effort:
\begin{equation}
    \begin{aligned}
    r_t^{reg} = & -\lambda_{ori} \left(1 - \left| \mathbf{q}_t^\top \mathbf{q}_{I} \right|\right) - \lambda_{\omega} \| \boldsymbol{\omega}_t \| \\
              & -\lambda_{yaw} |\psi_t| - \lambda_{a} \| \mathbf{a}_t \| - \lambda_{\Delta a} \| \mathbf{a}_t - \mathbf{a}_{t-1} \|
    \end{aligned}
\end{equation}
\noindent where $\mathbf{q}_I=[1,0,0,0]$ is the identity quaternion representing a level hover, $\psi_t$ is the yaw angle derived from the quaternion, and $\mathbf{a}_t$ is the action vector. 

All reward parameters are set as follows: $\lambda_{xy}=0.5$, $\lambda_z=0.2$, $\beta=0.5$, $v_{min}=0.01$ m/s, $v_{max}=0.5$ m/s, $\lambda_{vis}=0.1$, $\lambda_{ori}=0.005$, $\lambda_{\omega}=0.001$, $\lambda_{yaw}=0.002$, $\lambda_a=0.005$, and $\lambda_{\Delta a}=0.0025$.

\subsubsection{Domain Randomization and Curriculum Learning}
During training, we utilize landing platforms of fundamental geometric shapes, such as circles, cubes, and triangles, with inner diameters ranging from $0.2\text{m}$ to $0.5\text{m}$ and heights between $0.01\text{m}$ and $0.5\text{m}$ to provide semantic masks. We also randomize the initial positions of the quadrotor and the platform. For landing on moving platforms, the platform follows random trajectories sampled from a diverse motion set, including straight lines, cubic curves, circular arcs, and polygonal paths, with speeds varying from 0 to 2~m/s. To minimize the sim-to-real gap, we introduce noise to the centroid features extracted from the semantic masks. Specifically, small random offsets are added to the normalized pixel coordinates, and the visibility indicator is randomly flipped with a low probability to simulate segmentation errors and missed detections commonly encountered in real-world visual perception. 

We start training with static platforms and randomization to stabilize basic descent and visual alignment. We then gradually introduce dynamic platforms by first enabling low-speed motion, and progressively increasing motion complexity (cubic/circular/polygonal paths) and speed variability. This curriculum encourages the policy to first master stable static landing and visual tracking, and then become robust to unscripted time-varying dynamic landing.

\subsubsection{Network Architecture}
The mask is encoded by a three-layer CNN (channels: 6--12--18, kernels: 5--3--3), followed by a 256--256 MLP. The 13-dimensional state vector is encoded by a 192--192 MLP. The concatenated features are processed by a 192--96 actor MLP to produce actions. All hidden layers use ReLU activations.

\section{Experiments}
\subsection{Experimental Settings}
We employ the VisFly \cite{visfly} simulator, a fast-rendering simulation platform tailored for vision-based quadrotor tasks. The training process directly implements gradient backpropagation using the short-horizon BPTT algorithm \cite{shac}. All simulation and training processes are conducted on an Ubuntu 20.04 machine equipped with an Intel i9-13900K processor and an RTX-4090 GPU. To enhance the reproducibility of experiments, we provide the specific training parameters for the quadrotor platform and the algorithms in Tab.~\ref{tab:training_para}. We use the success rate (SR) and the landing error (LE) as evaluation metrics. A trial is successful when both the horizontal and vertical distances from the quadrotor to the platform center are below $0.20$~m. In the real-world experiments, this condition triggers motor-speed reduction, and the trial is successful only if the quadrotor remains stable on the platform after becoming stationary. LE is the horizontal error averaged over successful trials. In the ral world, LE is measured by Vicon.

\begin{table}[htbp]
\centering
\caption{Parameters.}
\label{tab:training_para}
\footnotesize
\setlength{\tabcolsep}{2.5pt}
\begin{tabular}{lccc}
\toprule
\textbf{Parameter} & \textbf{BPTT} & \textbf{PPO} & \textbf{SAC} \\
\midrule
Optimizer & Adam & Adam & Adam \\
LR schedule & $10^{-3}{\to}10^{-5}$ & $10^{-4}{\to}10^{-5}$ & $10^{-4}{\to}10^{-5}$ \\
Discount & 0.99 & 0.99 & 0.99 \\
Horizon & 96 & -- & -- \\
Clip range & -- & 0.2 & -- \\
GAE-$\lambda$ & -- & 0.95 & -- \\
Soft update $\tau$ & -- & -- & 0.005 \\
Replay size & 100000 & -- & 1000000 \\
Batch size & 25600 & 25600 & 512 \\
\midrule
\textbf{Quadrotor} & \multicolumn{3}{c}{\textbf{Value}} \\
Mass [kg] & \multicolumn{3}{c}{0.56} \\
Inertia [g$\cdot$m$^2$] & \multicolumn{3}{c}{$[1.25,1.28,2.03]$} \\
Max. thrust/motor [N] & \multicolumn{3}{c}{4.12} \\
Arm length [m] & \multicolumn{3}{c}{0.075} \\
Motor constant [s] & \multicolumn{3}{c}{0.035} \\
\bottomrule
\end{tabular}
\end{table}

\subsection{Baseline Comparison}
To evaluate the performance of our landing policy, we compare BPTT with PPO~\cite{ppo}, SAC~\cite{sac}, a centroid servo, and centroid-and-corner IBVS (CC-IBVS)~\cite{keipour2022visual}. Except for CC-IBVS, which additionally uses four bounding-box corners together with fixed camera calibration and nominal platform-size parameters, all methods use exactly the same inputs and CTBR outputs. All methods are evaluated on the same 50 static and 50 moving scenarios. For learning-based methods, SR and LE are averaged over three independently trained seeds, with LE computed over successful trials. All three learning-based methods use the same dense reward terms, while PPO and SAC additionally receive terminal success and failure rewards to facilitate model-free exploration. Each learning-based method is trained for 50 million environment steps using the same set of three random seeds. For each run, the final value is the mean episodic reward over the last 1,000 episodes, and convergence is the earliest point at which the 100-episode moving average remains within 5\% of this value for 50 consecutive episodes. Algorithm-specific hyperparameters are shown in Tab.~\ref{tab:training_para}. In addition to SR and LE, we also record their training time to reach convergence for learning-based methods.

\begin{figure}[htbp]
    \centering
    \includegraphics[width=0.68\linewidth]{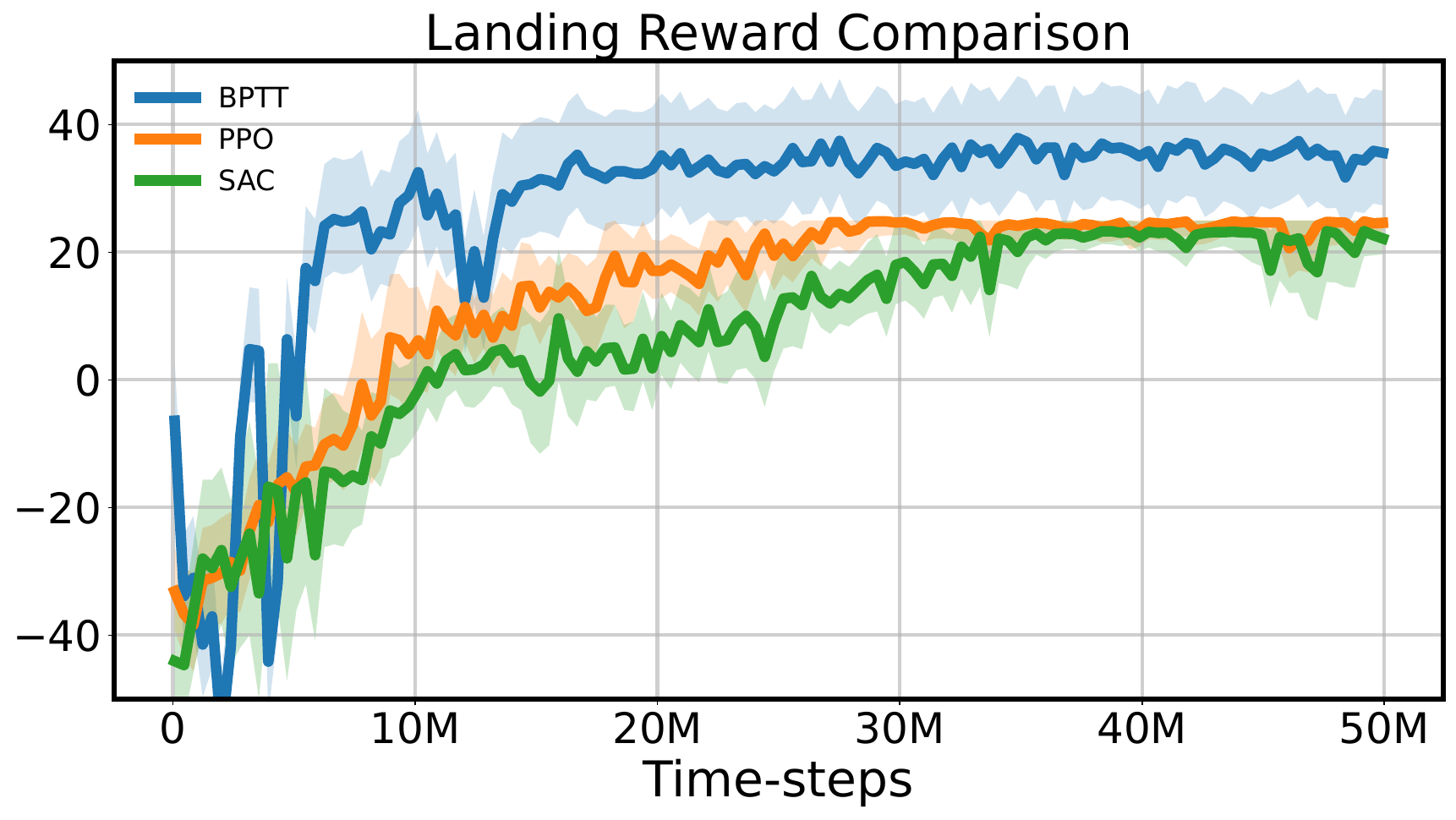}
    \caption{Learning curves of BPTT, PPO, and SAC under identical settings.}
    \label{fig:baselines}
\end{figure}

Tab.~\ref{tab:baselines} summarizes all baselines, while Fig.~\ref{fig:baselines} compares the learning curves. Our method benefits from direct optimization through analytic gradients, resulting in exceptionally high sample efficiency. Specifically, BPTT converges more than twice as fast as PPO and SAC because it backpropagates analytic gradients through differentiable dynamics. PPO and SAC instead treat the dynamics as a black box and rely on model-free interaction and exploration. Both visual-servoing methods handle static platforms but degrade markedly on moving targets, reflecting the limitations of fixed servo laws for coupled tracking and descent. Our policy achieves the highest SR and remains robust on moving targets. These results clearly demonstrate that differentiable physics learning provides a much more sample-efficient and effective paradigm for learning stable and adaptive landing policies.

\begin{table}[htbp]
\centering
\caption{Baseline comparison (SR/LE in \%/m).}
\label{tab:baselines}
\begin{tabular}{lccc}
\toprule
\textbf{Method} & \textbf{Static SR/LE} & \textbf{Moving SR/LE} & \textbf{Conv. [h]} \\
\midrule
PPO & {94/0.07} & {70/0.15} & 5.2 \\
SAC & {90/0.08} & {52/0.16} & 5.7 \\
Centroid servo & 82/0.09 & 48/0.20 & -- \\
CC-IBVS~\cite{keipour2022visual} & 96/0.07 & 64/0.18 & -- \\
\textbf{Ours (BPTT)} & {\textbf{98/0.06}} & {\textbf{92/0.12}} & \textbf{2.5} \\
\bottomrule
\end{tabular}
\end{table}

\subsection{Ablation Study} 
We conduct ablative studies to validate the design of our strategy, specifically focusing on four key components: the effectiveness of the adaptive descent reward $r_t^{vz}$, the role of the visual alignment reward $r_t^{vis}$, the contribution of semantic segmentation masks versus centroid-only features and centroid features augmented with normalized mask area, and the dynamic platforms curriculum. All ablation experiments are conducted with identical experimental settings and evaluated over 100 randomly generated static and dynamic landing platforms.

\begin{figure}[htbp]
    \centering
    \subfloat[{\footnotesize Ours (full) with $r_t^{vz}$.}\label{fig:vz_curve_full}]{\includegraphics[width=0.7\linewidth]{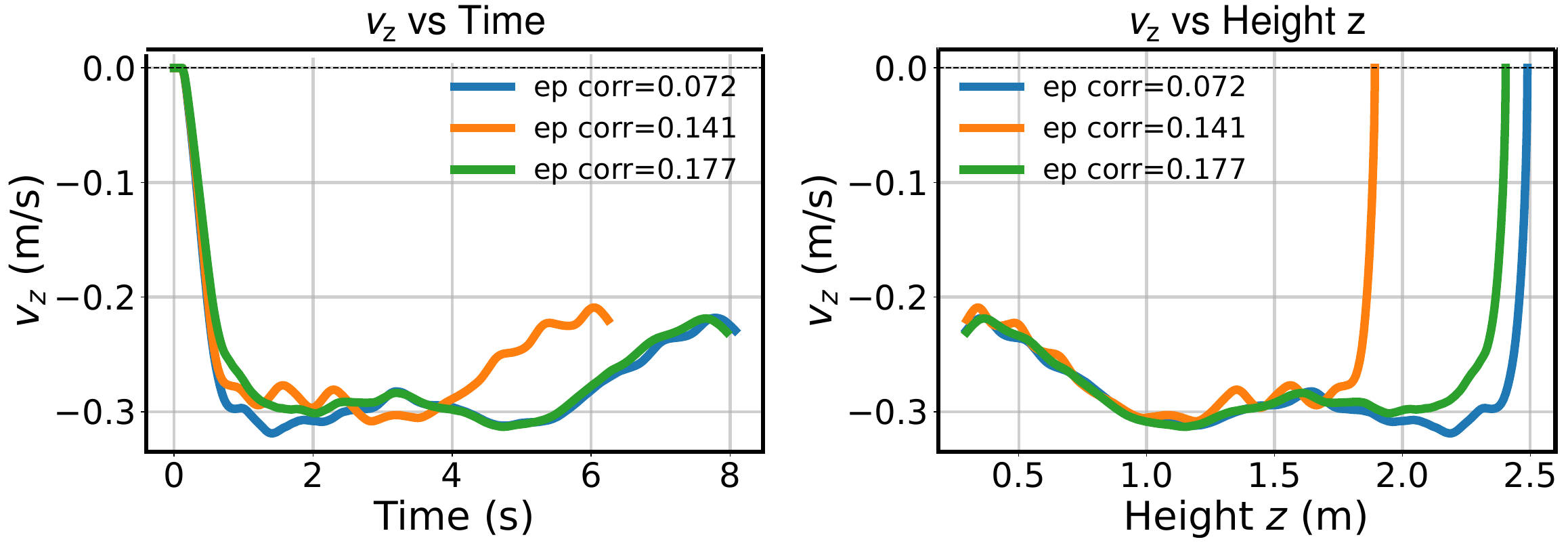}}\\[0.15em]
    \subfloat[{\footnotesize Ablation study: w/o $r_t^{vz}$.}\label{fig:vz_curve_wo}]{\includegraphics[width=0.7\linewidth]{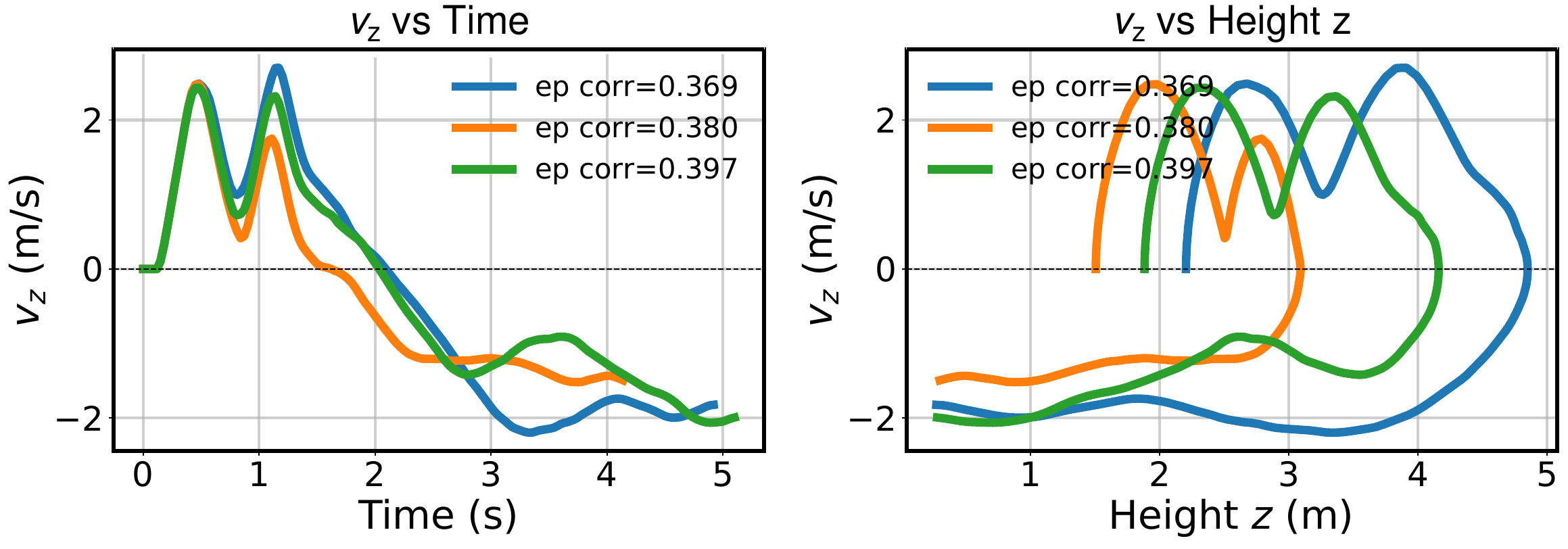}}
    \caption{Descent velocity with and without $r_t^{vz}$; ep\_corr is the altitude--velocity correlation (lower is better).}
    \label{fig:vz_curve}
\end{figure}

Tab.~\ref{tab:ablation} presents the SR and LE of the ablation study. It can be observed that the performance of our policy deteriorates when any one of the four components is ablated. Although without $r_t^{vz}$ the policy still achieves relatively high SR and low LE, it cannot adapt descent velocity to altitude. Fig.~\ref{fig:vz_curve} shows three landing trajectories starting from random positions for each strategy. As shown in Fig.~\ref{fig:vz_curve_full}, when $r_t^{vz}$ is included, the descent speed adaptively adjusts according to the altitude, exhibiting a stable landing process characterized by initially rapid descent followed by slow velocity near touchdown, which ensures safety at landing. In the absence of $r_t^{vz}$ (Fig.~\ref{fig:vz_curve_wo}), the descent process becomes highly unstable. At the moment of touchdown, the descent velocity spikes to approximately $-2\,\text{m/s}$, resulting in a highly dangerous landing that lacks smoothness and stability.

In addition, removing the visual alignment reward $r_t^{vis}$ causes the SR to drop sharply to 22\%, with a marked increase in LE, indicating that the model frequently loses target tracking when there is no prior information about the target position and no explicit incentive to keep the target centered in the observation. Centroid-only remains effective (78\% SR, 0.13~m LE), confirming that the centroid provides an image-plane alignment cue. Adding normalized mask area improves performance by providing an apparent-scale cue. However, both compress the target into a few scalars. The full mask reaches 99\% SR and 0.08~m LE because it preserves spatial extent and contour, providing shape, perspective, and partial-visibility cues without explicit depth estimation or a predefined platform size. Due to the visual alignment reward, the policy automatically aligns with the platform, which itself provides generalization to minor platform motions. However, without the dynamic platforms curriculum, the policy struggles to generalize to irregularly moving dynamic targets, resulting in a significant drop in performance. The ablation study validates the importance of the adaptive descent reward, the visual alignment reward, the semantic segmentation representation, and the dynamic platforms curriculum for robust and generalizable landing.

\begin{table}[htbp]
\centering
\caption{Ablation study.}
\label{tab:ablation}
\begin{tabular}{lcc}
\toprule
\textbf{Ablation} & \textbf{SR [\%]} & \textbf{LE [m]} \\
\midrule
w/o $r_t^{vz}$ (adaptive descent) & 90\% & 0.12 \\
w/o $r_t^{vis}$ (visual alignment) & 22\% & 0.18 \\
w/o dynamic platforms curriculum & 45\% & 0.20 \\
Centroid-only (no segmentation mask) & 78\% & 0.13 \\
Centroid + normalized mask area & 91\% & 0.12 \\
\textbf{Ours (full)} & \textbf{99\%} & \textbf{0.08} \\
\bottomrule
\end{tabular}
\begin{minipage}{\linewidth}\scriptsize\raggedright
Tables~II and~III use different scenarios. Therefore, the SR are not directly comparable.
\end{minipage}
\end{table}

\subsection{Sim-to-real Transfer}
\begin{figure}[htbp]
    \centering
    \includegraphics[width=0.70\linewidth]{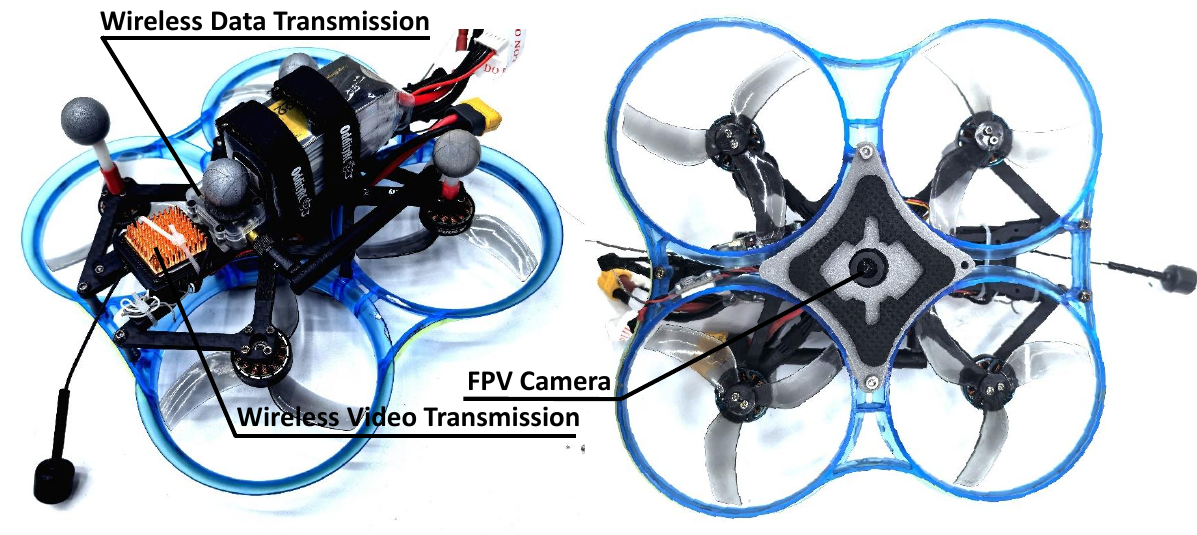}
    \caption{Experimental quadrotor with a downward-facing monocular camera.}
    \label{fig:drone}
\end{figure}

We deploy the learned policy on a real quadrotor, which is equipped with a data-link module, a video-transmission module, and a downward-facing monocular camera for offboard computation, as shown in Fig.~\ref{fig:drone}. Inferences for both the policy and the open-vocabulary visual frontend are executed on a laptop equipped with an NVIDIA RTX 5090 GPU. This setup is computationally demanding because generating the segmentation mask requires running both YOLO-World and MobileSAM in real time. The output frequency of the semantic segmentation mask and the control frequency are both $30\,\text{Hz}$. The linear velocity is provided by the indoor Vicon motion capture system, and the attitude and angular velocity are provided by the onboard IMU. The ground-station policy denormalizes its CTBR output and publishes collective-thrust acceleration and body-rate commands through ROS. The interface uses a calibrated thrust model to map acceleration to normalized throttle and sends it with the body rates as MAVLink \texttt{ATTITUDE\_TARGET} messages to an F722 Betaflight controller. Its onboard rate loop and mixer then generate individual motor commands. 

\begin{figure*}[htbp]
    \centering
    \includegraphics[width=0.68\linewidth]{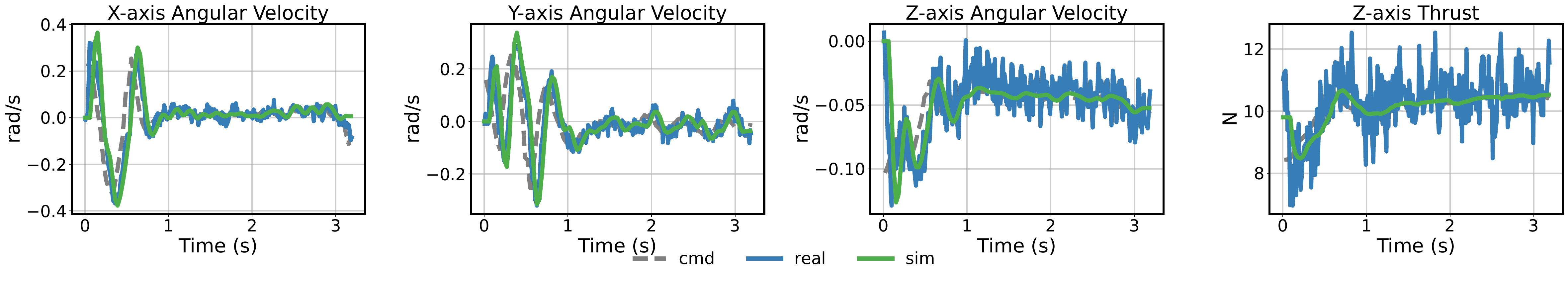} 
    \caption{Simulated and measured CTBR responses after incorporating the measured $120\,\text{ms}$ round-trip delay.}
    \label{fig:delay_alignment}
\end{figure*}

Before the real-world deployment, we conduct system identification to measure the total round-trip delay of approximately $120\,\text{ms}$ caused by data and image transmission, and control command uplink. By explicitly compensating for this delay in simulation training, our policy becomes aware of and adapts to the real-world latency. After delay compensation, the simulated and real-world CTBR response curves are highly aligned (see Fig.~\ref{fig:delay_alignment}), effectively narrowing the sim-to-real gap.

\subsubsection{Landing on Static Platforms}
\begin{figure}[htbp]
    \centering
    \includegraphics[width=0.72\linewidth]{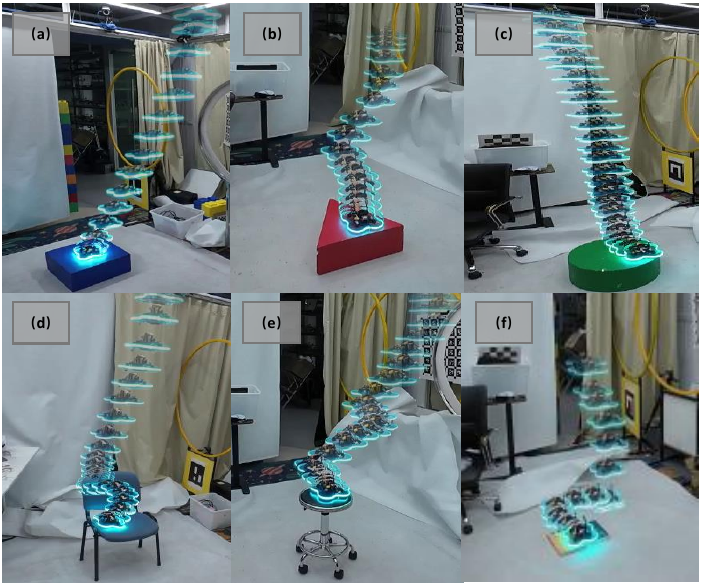}
    \caption{Real-world trajectories on seen (a--c) and unseen (d--f) platforms.}
    \label{fig:static_landing}
\end{figure}

To validate the open-vocabulary landing capability of the system in real-world settings, we first deploy the quadrotor on diverse static platforms, using natural language prompts to specify landing targets. The system switches among different platforms by changing only the text prompt supplied to the frozen perception frontend. As seen in Fig.~\ref{fig:static_landing} and Tab.~\ref{tab:sim2real_platform_trials}, our approach enables robust and stable landings across various platform types, including both shapes encountered during simulation (circle, rectangle, triangle) and novel, previously unseen objects (e.g., books, chairs, suitcase). For quantitative real-world evaluation, we conduct 10 real-world landing trials for each platform type (both seen and unseen), with random initial positions. Tab.~\ref{tab:sim2real_platform_trials} lists the platform snapshots, trial-level outcomes (Trials/Succ.), and aggregate SR and LE for both seen and unseen categories at the bottom. The controller achieves 96.7\% SR with low LE on seen geometries and 80.0\% on targets excluded from controller training.

For unsuccessful trials in Tab.~\ref{tab:sim2real_platform_trials}, most cases stem from open-vocabulary mask errors, including missed detections, abruptly flickering segmentation, or excessively small in-view targets (e.g., books), which bias centroid estimates and induce lateral error. The remaining failures often involve physically compact landing platforms (e.g., the triangular pad or the black stool). After touchdown, a center-of-mass offset relative to the support causes the quadrotor to tip and fall.

\begin{table}[htbp]
    \begingroup
    \centering
    \caption{Real-world landing performance on different platforms.}
    \label{tab:sim2real_platform_trials}
    \footnotesize
    \setlength{\tabcolsep}{3.2pt}%
    \renewcommand{\arraystretch}{1.0}%
    \def\simpT{\dimexpr\columnwidth-10\tabcolsep-9pt\relax}%
    \def\simpImgW{\dimexpr\simpT*22/100\relax}%
    \def\simpRateW{\dimexpr\simpT*11/100\relax}%
    \def\trialsuccraise{2.0ex}%
    \def\ratemid#1{\raisebox{-\trialsuccraise}[0.75ex][0.75ex]{\makecell[c]{#1}}}%
    \def\simpRateTxt{\centering\arraybackslash}
    \def\platimg#1#2#3{%
        \begin{minipage}[c]{\linewidth}\centering
        #1\\[-1pt]
        {\scriptsize #2}%
        \end{minipage} & \multicolumn{1}{>{\simpRateTxt}m{\simpRateW}}{\ratemid{#3}}}%
    \begin{tabular}{@{}>{\centering\arraybackslash}m{\simpImgW}>{\simpRateTxt}m{\simpRateW}>{\centering\arraybackslash}m{\simpImgW}>{\simpRateTxt}m{\simpRateW}>{\centering\arraybackslash}m{\simpImgW}>{\simpRateTxt}m{\simpRateW}@{}}
        \toprule
        \multicolumn{6}{c}{\textbf{Different types of platforms (Trials/Succ.)}} \\
        \midrule
        \platimg{\includegraphics[width=\linewidth,height=0.65cm,keepaspectratio]{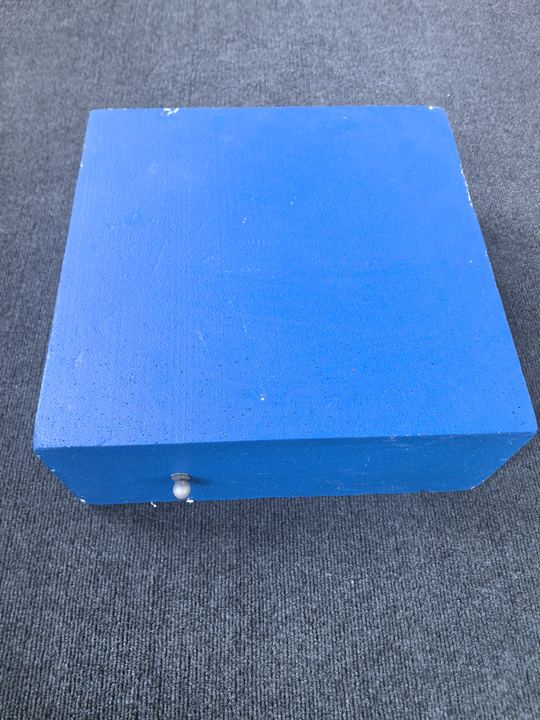}}{$0.5{\times}0.5\,\mathrm{m}$}{$10/10$} &
        \platimg{\includegraphics[width=\linewidth,height=0.65cm,keepaspectratio]{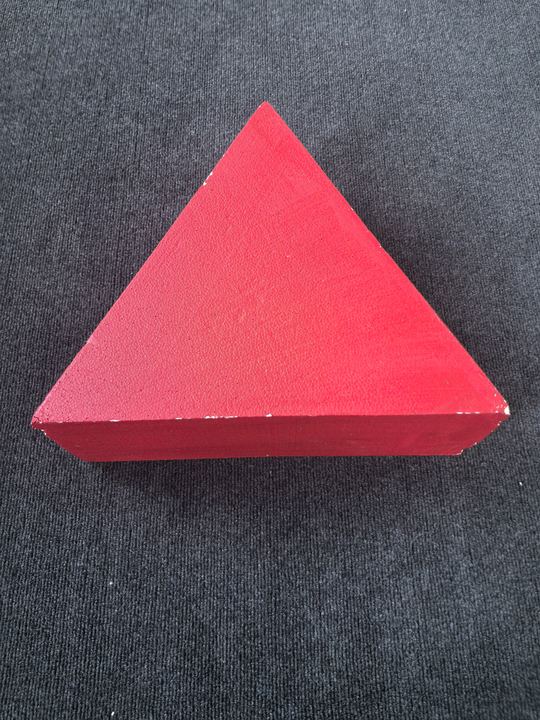}}{$R{=}0.25\,\mathrm{m}$}{$10/9$} &
        \platimg{\includegraphics[width=\linewidth,height=0.65cm,keepaspectratio]{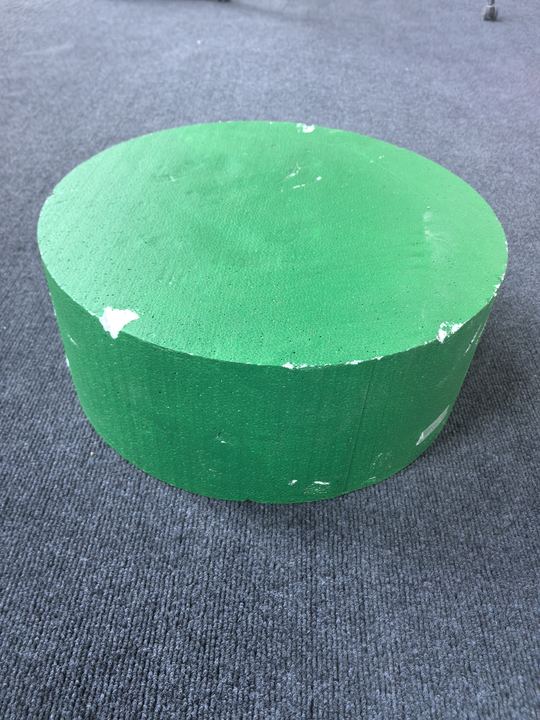}}{$R{=}0.25\,\mathrm{m}$}{$10/10$} \\
        \midrule
        \platimg{\includegraphics[width=\linewidth,height=0.65cm,keepaspectratio]{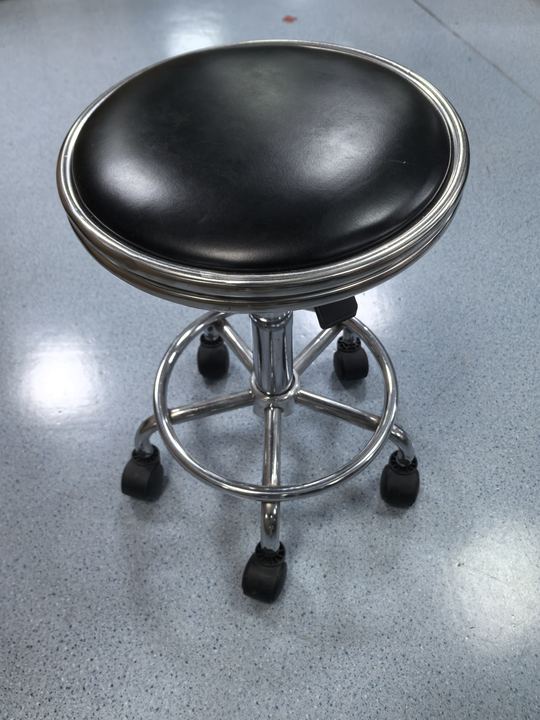}}{$R{=}0.16\,\mathrm{m}$}{$10/8$} &
        \platimg{\includegraphics[width=\linewidth,height=0.65cm,keepaspectratio]{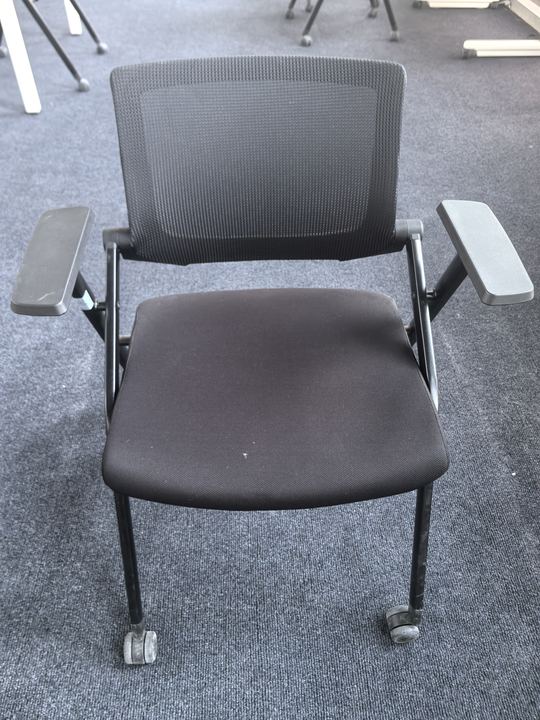}}{$0.45{\times}0.4\,\mathrm{m}$}{$10/8$} &
        \platimg{\includegraphics[width=\linewidth,height=0.65cm,keepaspectratio]{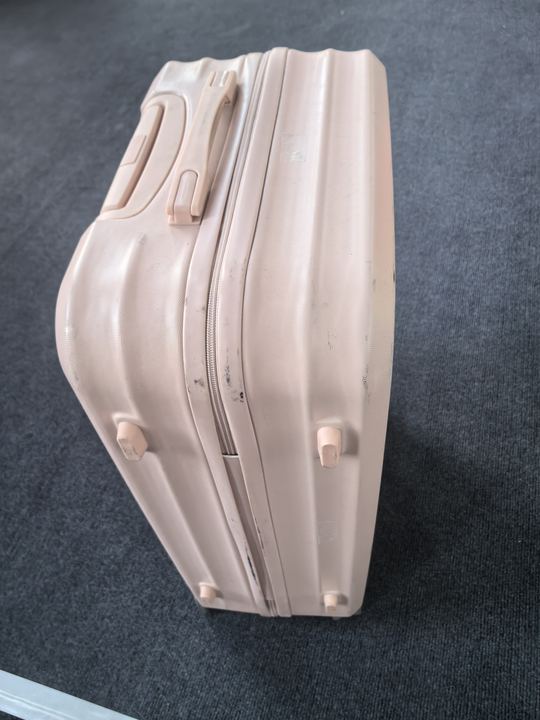}}{$0.69{\times}0.4\,\mathrm{m}$}{$10/10$} \\
        \midrule
        \platimg{\includegraphics[width=\linewidth,height=0.65cm,keepaspectratio]{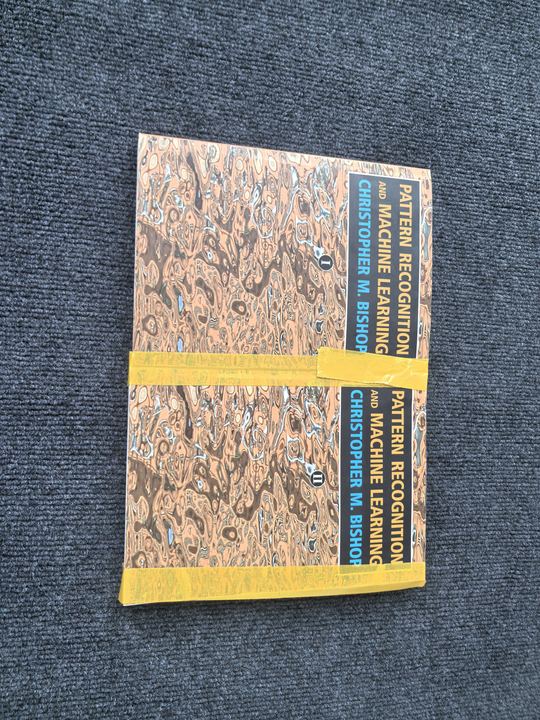}}{$0.4{\times}0.28\,\mathrm{m}$}{$10/8$} &
        \platimg{\includegraphics[width=\linewidth,height=0.65cm,keepaspectratio]{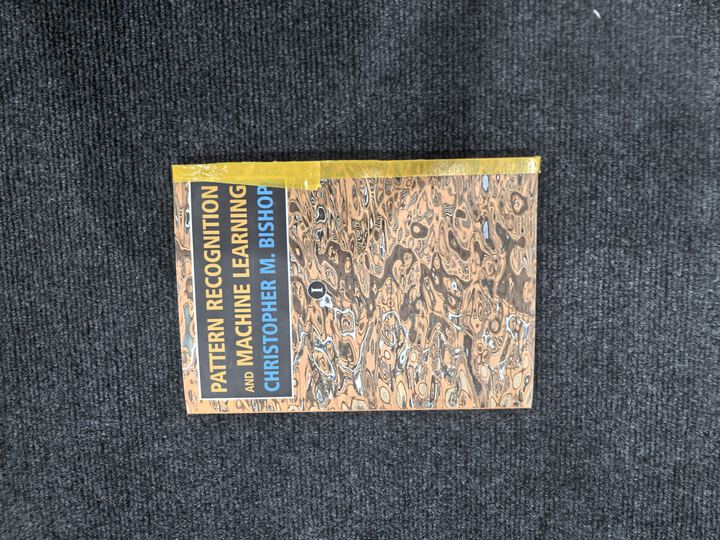}}{$0.2{\times}0.28\,\mathrm{m}$}{$10/6$} &
        \platimg{\includegraphics[width=\linewidth,height=0.65cm,keepaspectratio]{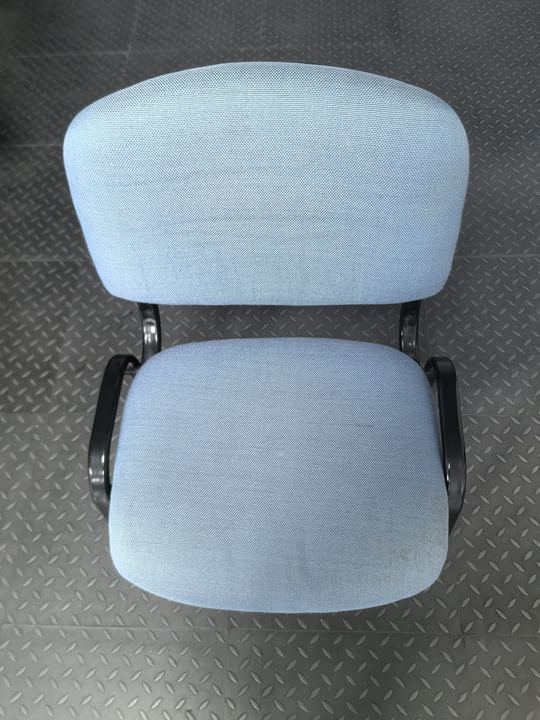}}{$0.46{\times}0.41\,\mathrm{m}$}{$10/8$} \\
        \midrule
        \multicolumn{4}{@{}l@{}}{\textbf{Platform type}} & \multicolumn{1}{>{\centering\arraybackslash}m{\simpImgW}}{\textbf{SR[\%]}} & \multicolumn{1}{>{\simpRateTxt}m{\simpRateW}}{\textbf{LE[m]}} \\
        \midrule
        \multicolumn{4}{l}{Seen (circle/cube/triangle)} & \multicolumn{1}{>{\centering\arraybackslash}m{\simpImgW}}{\makecell[c]{96.7\%}} & \multicolumn{1}{>{\simpRateTxt}m{\simpRateW}}{\makecell[c]{0.09}} \\
        \multicolumn{4}{l}{Unseen (e.g., books, chairs, suitcase)} & \multicolumn{1}{>{\centering\arraybackslash}m{\simpImgW}}{\makecell[c]{80.0\%}} & \multicolumn{1}{>{\simpRateTxt}m{\simpRateW}}{\makecell[c]{0.13}} \\
        \bottomrule
    \end{tabular}\\[2pt]
    \begin{minipage}{\linewidth}\scriptsize\raggedright
    $R$ denotes the radius of the platforms.
    \end{minipage}
    \endgroup
\end{table}

\subsubsection{Landing on Moving Platforms}
\begin{figure}[htbp]
    \centering
    \includegraphics[width=0.72\linewidth]{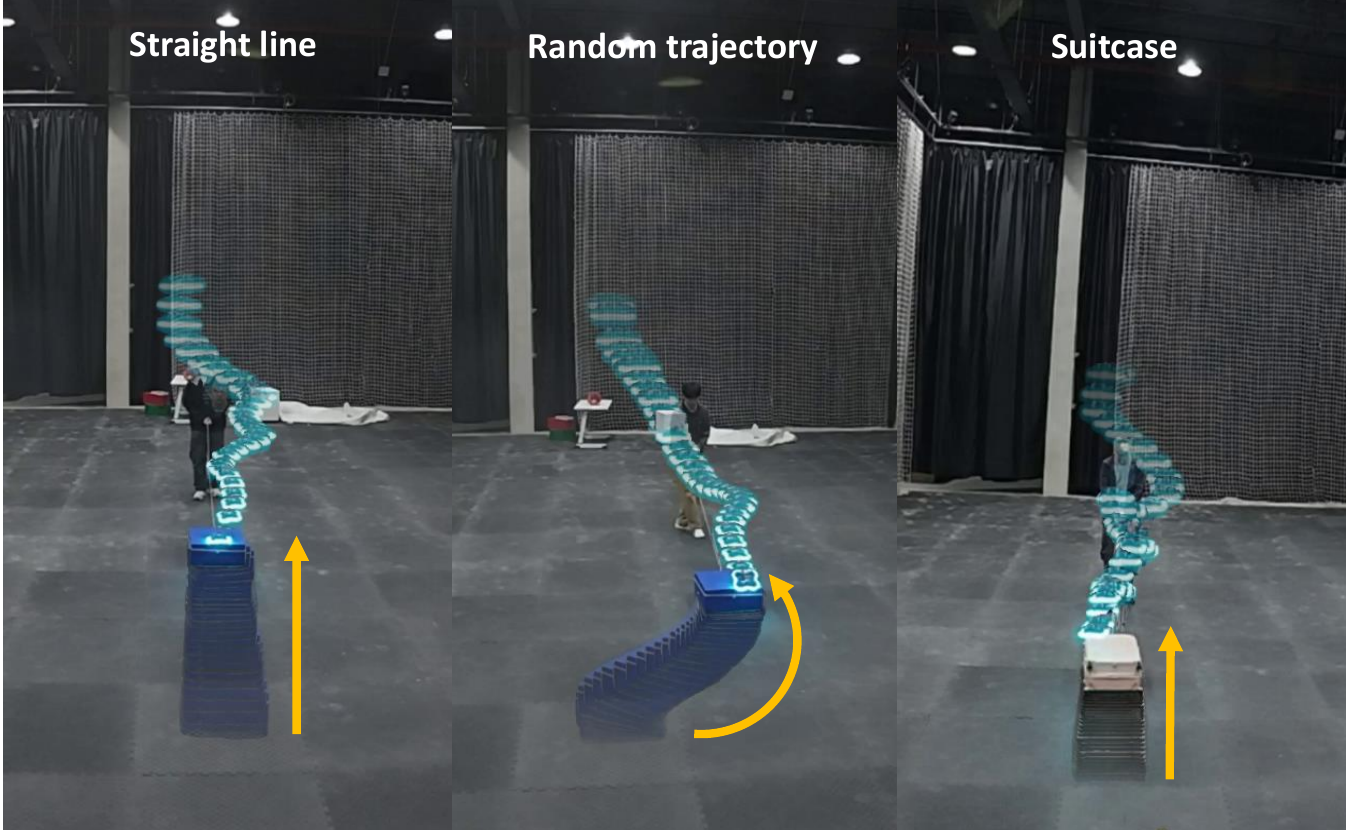}
    \caption{Real-world landing trajectories for a straight path, a semicircular path, and a moving suitcase (left to right).}
    \label{fig:real_world_dynamic}%
\end{figure}

We further evaluate the landing policy on dynamic platforms, where the platform moves on the ground during descent. To produce random platform trajectories, we manually pull the platform during dynamic landing tests. For quantitative evaluation, we control the platform's ground-plane speed during each trial and group the trials into three speed ranges: $[0, 0.5)\,\text{m/s}$, $[0.5, 1)\,\text{m/s}$, and $[1, 2]\,\text{m/s}$. Within each range, we conduct 10 real-world landing trials with random initial positions of both the quadrotor and the platform. Tab.~\ref{tab:dynamic_real_world} summarizes SR, LE, Vicon ground-truth platform speed and acceleration, and path curvature computed from the recorded horizontal trajectories for each group of trials, and Fig.~\ref{fig:real_world_dynamic} shows the real-world dynamic landing visualizations on both seen and unseen platforms with different trajectories. 

The results in Tab.~\ref{tab:dynamic_real_world} show that our policy exhibits strong generalization to dynamic platforms with random low-speed ground motion along diverse trajectories. Fig.~\ref{fig:real_world_dynamic} further demonstrates stable landings on both seen and unseen moving platforms when the target follows diverse trajectories, together with exceptionally smooth quadrotor landing trajectories.

\begin{table}[htbp]
    \centering
    \caption{Real-world dynamic landing performance.}
    \label{tab:dynamic_real_world}
    \footnotesize
    \setlength{\tabcolsep}{2.6pt}
    \renewcommand{\arraystretch}{0.9}
    \begin{tabular}{@{}lccccc@{}}
    \toprule
    \makecell{\textbf{Speed bin}\\\textbf{[m/s]}} &
    \textbf{SR [\%]} &
    \textbf{LE [m]} &
    \makecell{\textbf{Mean speed}\\\textbf{[m/s]}} &
    \makecell{\textbf{Mean accel.}\\\textbf{[m/s$^2$]}} &
    \makecell{\boldmath$\kappa_{50}/\kappa_{95}$\\\textbf{[m$^{-1}$]}} \\
    \midrule
    $[0,0.5)$ & 100 & 0.11 & $0.38\pm0.06$ & $0.28\pm0.13$ & $0.08/1.23$ \\
    $[0.5,1)$ & 100 & 0.14 & $0.72\pm0.12$ & $0.51\pm0.11$ & $0.03/0.95$ \\
    $[1,2]$ & 90 & 0.16 & $1.33\pm0.21$ & $0.84\pm0.31$ & $0.04/0.82$ \\
    \bottomrule
    \end{tabular}
    \\[2pt]
    \begin{minipage}{\columnwidth}\scriptsize
    Vicon speed and acceleration are reported as mean$\pm$SD over ten trials. $\kappa_{50}/\kappa_{95}$ are the median/P95 horizontal-trajectory curvatures.
    \end{minipage}
\end{table}

\FloatBarrier
\section{Conclusion and Limitation}
We propose \textit{Ask to Land}, a mask-conditioned autonomous drone landing framework that combines a frozen open-vocabulary perception frontend for text-prompt-based target recognition with policy optimization via BPTT in a differentiable quadrotor simulator. As a result, the learned policy can adapt the descent behavior without handcrafted heuristics, supporting both stable static landing and robust dynamic landing. Experimental results demonstrate that the proposed framework achieves consistently high success rates and low landing errors without target-specific fine-tuning. In addition to these contributions, our approach has not yet addressed scenarios with highly cluttered backgrounds, strong outdoor wind disturbances, or high-speed dynamic tracking. Furthermore, due to current computational resource limitations, direct deployment on onboard computing units remains challenging. Future work will focus on advancing universal landing capabilities for outdoor platforms with high-speed tracking and complex backgrounds, and on enabling efficient real-time inference on embedded hardware.

\bibliographystyle{IEEEtran}
\bibliography{references.bib}

\end{document}